%% file: colm2024_conference.tex
\documentclass{article} 
\usepackage{colm2024_conference}

\usepackage{microtype}
\usepackage{hyperref}
\usepackage{url}
\usepackage{booktabs}
\usepackage{graphicx}%
\usepackage{multirow}%
\usepackage{amsmath,amssymb,amsfonts}%
\usepackage{amsthm}%
\usepackage{mathrsfs}%
\usepackage[title]{appendix}%
\usepackage{xcolor}%
\usepackage{textcomp}%
\usepackage{manyfoot}%
\usepackage{booktabs}%
\usepackage{algorithm}%
\usepackage{algorithmicx}%
\usepackage{algpseudocode}%
\usepackage{listings}%

\title{MING-MOE: Enhancing Medical Multi-Task Learning in Large Language Models with Sparse Mixture of Low-Rank Adapter Experts}

\colmfinalcopy

\author{Yusheng Liao\thanks{Equal Contribution} \\
Cooperative Medianet Innovation Center,\\Shanghai Jiao Tong University \\
\texttt{liao20160907@sjtu.edu.cn} 
\And
Shuyang Jiang$^*$ \\
Fudan University\\
\texttt{shuyangjiang23@m.fudan.edu.cn} \\
\AND
Yu Wang\thanks{Corresponding Author}, Yanfeng Wang \\
Cooperative Medianet Innovation Center, Shanghai Jiao Tong University \\
Shanghai Artificial Intelligence Laboratory \\
\texttt{\{yuwangsjtu, wangyanfeng622\}@sjtu.edu.cn} \\
}

\begin{document}

\maketitle

\begin{abstract}
Large language models like ChatGPT have shown substantial progress in natural language understanding and generation, proving valuable across various disciplines, including the medical field. Despite advancements, challenges persist due to the complexity and diversity inherent in medical tasks which often require multi-task learning capabilities. Previous approaches, although beneficial, fall short in real-world applications because they necessitate task-specific annotations at inference time, limiting broader generalization. This paper introduces MING-MOE, a novel Mixture-of-Expert~(MOE)-based medical large language model designed to manage diverse and complex medical tasks without requiring task-specific annotations, thus enhancing its usability across extensive datasets. MING-MOE employs a Mixture of Low-Rank Adaptation (MoLoRA) technique, allowing for efficient parameter usage by maintaining base model parameters static while adapting through a minimal set of trainable parameters. We demonstrate that MING-MOE achieves state-of-the-art (SOTA) performance on over 20 medical tasks, illustrating a significant improvement over existing models. This approach not only extends the capabilities of medical language models but also improves inference efficiency.
\end{abstract}
\input{1_intro}

\input{2_related}
\input{3_method}
\input{4_exp}
\input{5_discussion}
\input{6_conclusion}


\bibliography{colm2024_conference}
\bibliographystyle{colm2024_conference}


\end{document}

%% file: 1_intro.tex
\section{Introduction}
Large language models such as ChatGPT~\citep{elmohamed} have made significant strides in the realms of natural language understanding and generation, marking a pivotal advancement in the field of artificial intelligence. These models have demonstrated exceptional capabilities in following instructions and completing tasks, thereby finding extensive applications across a broad spectrum of disciplines~\citep{LAWGPT-zh, deng2024k2}. 

Among these, the medical domain has garnered considerable attention due to its significance and demand, with a substantial amount of work focused on researching and developing specialized large language models~\citep{li2023chatdoctor,wang2023huatuo}. Though progress has been made through these efforts, the large language models still face the challenges of complexity and diversity in medical tasks, where the data is often deliberately structured to represent a diverse set of tasks, also known as multi-task learning~\citep{chung2022scaling, wei2021finetuned, sanh2021multitask}. Specifically, medical tasks such as medical question answering~\citep{zhang2018medical}, medical dialogues~\citep{liu2022meddg,zhao2022medical}, medical entity recognition~\citep{hongying2021building} and clinical terminology standardization~\citep{zhang2022cblue} have significant differences in the formats of input and output sequences, which make it difficult for large language models to learn and process them simultaneously. 

Fortunately, several studies have explored Mixture of Experts (MoE)-based methods in the general domain, demonstrating its potential to surmount the multi-task learning challenge~\citep{zadouri2023pushing}. Unlike traditional models that utilize all parameters for every input, the MoE structure comprises several sub-modules, also known as experts, employed to process different inputs individually~\citep{JMLR:v23:21-0998}.  The MoE-based models introduce separate experts to learn from different tasks, and automatically call affinity experts for a specific downstream task, which avoids task-aliasing and enhances models' performance in multi-task learning.
Selectively employing expert modules for various tasks can also improve the model's efficiency during inference~\citep{du2022glam}. Moreover, some experiments find that the MOE-based structures can nearly fully efficient in storing knowledge~\citep{allenzhu2024physics}. In medical applications, prior studies have employed a task-specific Mixture of Experts~(MoE)-based parameter efficient fine-tuning~(PEFT) to augment the capacity of large language models~\citep{liu2023moelora}, allowing for the learning of distinct sets of parameters for each task. 
However, this method requires the specification of the task type for samples at test time, which is not suitable for real-world deployment applications and also limits the model's ability to further generalize to other unseen tasks. 

In this paper, we introduce MING-MOE~\footnote{Our code can be found in \url{https://github.com/MediaBrain-SJTU/MING}.}, an MoE-based medical large language model capable of executing complex and diverse medical tasks while maintaining state-of-the-art medical knowledge performance and conversational interaction capabilities. Unlike \citet{liu2023moelora}, MING-MOE can adaptively select the appropriate experts for each token during the training process instead of for each task. 
This token-level MoE architecture eliminates the need for task labels, thereby enabling its application to extensive datasets and improving its applicability and usability. 
Considering the impracticality of directly expanding a base model into a MoE architecture due to the exponential increase in parameters over the base model, we utilize a Mixture of Low-Rank Adaptation~(LoRA)~\citep{hu2021lora} approach to extend the model. 
This method freezes the base model's inherent parameters during training, requiring only a minimal number of parameters to be trained to fine-tune the MOE model, thus reducing the adaptation cost. 
We evaluate the capabilities of the medical large language models across two dimensions, including medical natural language processing~(NLP) tasks and medical licensing exams. 
The experiments are conducted on more than 20 medical tasks to comprehensively assess the performance of the MING-MOE. The results show that MING-MOE achieves the SOTA performance in currently available open-source medical models on all tasks.

In summary, our contributions are as follows:
\begin{itemize}
    \item We proposed a novel MOE-based medical large language model, MING-MOE, which achieves SOTA performance in medical multi-task learning. To the best of our knowledge, MING-MOE is the first MOE model capable of accomplishing such a diverse range of medical tasks.
    \item We conduct comprehensive experiments and compare the performance of MING-MOE with other models on more than 20 medical tasks, the results underscoring the superiority of the proposed MING-MOE models.
\end{itemize}

%% file: 2_related.tex
\section{Related Works}
\paragraph{Bilingual medical large language models}
Large language models such as GPT-4~\citep{elmohamed}, PaLM~\citep{chowdhery2023palm} and LLaMA~\citep{touvron2023llama,touvron2023llama2} have achieved superior zero-shot performance across tasks and serve as interactive chatbots to interact with humans. 
However, trained on little medical-oriented data and Chinese data, these LLMs are not useful for medical conversations and consultations, especially for Chinese medical scenarios.
Therefore, a lot of work has been done to fine-tune base models on medical data to obtain large medical models.
Med-PaLM~\citep{singhal2022large} is grounded on Flan-PaLM~\citep{chung2022scaling,chowdhery2023palm} to encode clinical knowledge.
Med-PaLM2~\citep{singhal2023towards} as a successor, reduce the gap with doctors by fine-tuning in the medical data and using modern prompting strategies.
Apart from grounding on super-large language models such as PaLM or GPT, many works attempt to build medical-LLM on deployable sizes of LLMs, including 7B and 13B. 
DoctorGLM~\citep{xiong2023doctorglm}, ChatGLM-Med~\citep{ChatGLM-Med}, and Bianque-2~\citep{chen2023bianque} are all built on ChatGLM to support acceptable bilingual medical consultation.
Other work like MedicalGPT~\citep{MedicalGPT}, Huatuo~\citep{wang2023huatuo} and HuatuoGPT-\uppercase\expandafter{\romannumeral2}~\citep{chen2023huatuogpt} are built on LLaMA-series and enlarge the vocabulary to support Chinese conversations.


\paragraph{Parameter efficient fine-uuning}

Full fine-tuning effectively adapts base large language models to downstream tasks but also consumes significant computational resources with the increasing size of models and the number of tasks. To address this, Parameter-Efficient Fine-Tuning (PEFT) methods have been introduced. These methods freeze the base language models, modifying only a negligible number of parameters during the training phase, yet achieving similar or even superior performance with limited fine-tuning data.
Among these methods, Adapter-Tuning~\citep{rebuffi2017learning,houlsby2019parameter,lin-etal-2020-exploring,pfeiffer-etal-2021-adapterfusion} was the pioneering architecture that connected two additional projection layers to the pretrained language model. 
In addition to incorporating additional modules, Prefix-tuning~\citep{li-liang-2021-prefix} introduces learnable prefix tokens and prepends them before the input prompts. 
Differently, Prompt-Tuning~\citep{lester-etal-2021-power} utilizes learnable prompt tokens for each task, as a multi-task PEFT approach in NLU scenarios. 
Following these, P-Tuning~\citep{LIU2023} and P-Tuning-v2~\citep{liu-etal-2022-p} move away from explicit prompts and employ a prompt-generator to convert pseudo prompts into task prompts, allowing decoder-only models to also perform NLU tasks. 
However, these methods introduce additional priors and significant inference latency.
In contrast, Low-Rank Adaptation (LoRA)\citep{hu2022lora} and its variant, Weight-Decomposed Low-Rank Adaptation (DoRA)\citep{liu2024dora}, take a different approach. 
LoRA updates original parameters with two low-rank matrices without assuming any specific task or architecture, eliminating inference latency by merging back these two matrices to the original weight. 
DoRA extends this by incorporating weight decomposition, achieving performance comparable to full fine-tuning. 
Nonetheless, transferring these methods to multi-task learning scenarios without manual adjustments remains a challenge.



%% file: 3_method.tex
\begin{figure}[tbp]
    \centering
    \includegraphics[width=\linewidth]{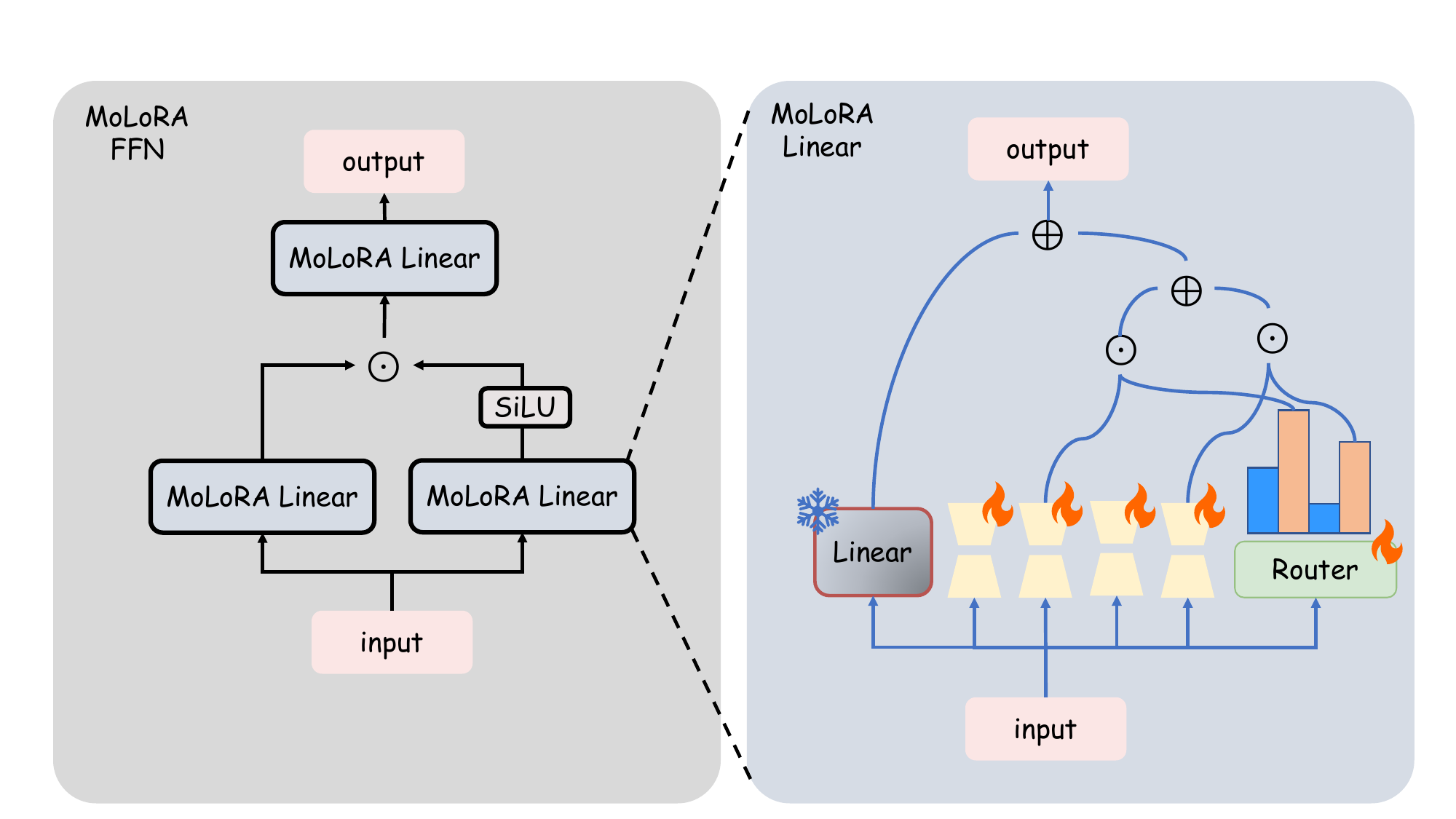}
    \caption{The MoLoRA FFN and corresponding MoLoRA Linear architecture in MING-MOE. This diagram shows a setting of $N=4$ experts and $K=2$ activated experts during training. }
    \label{fig:molora_linear}
\end{figure}
\section{MING-MOE}
In this section, we provide a comprehensive description
of our proposed framework.
We first describe the low-rank adaptation fine-tuning methods and then introduce our core architecture Mixture of Low-Rank Adaptation in MING-MOE.

\subsection{Preliminaries of Low-Rank Adaptation}
LoRA proves that the update of original linear layers $\Delta W$ in large language models is of low-rank, and can be decomposed into the multiplication of two compact matrices $BA$.
The pretrained weight $W$ is frozen in the training phase and does not undergo gradient updates, while $A$ and $B$ contain trainable parameters and contribute together to the forward pass:
\begin{equation}
\label{lora}
    h=Wx+\Delta W x = Wx + BAx
\end{equation}
where $W\in\mathbb{R}^{d\times d'},A\in\mathbb{R}^{r\times d},B\in\mathbb{R}^{d'\times r},x\in\mathbf{R}^{d}$ and $r\ll d,d'$.
At the beginning of training, $B$ is initialized to an all-zero matrix and $A$ uses Gaussian initialization to make sure that the product $BA$ is zero at initialization.

\subsection{Mixture of Low-rank Adaptation}
We base our MoE architecture on the Feed-forward Network~(FFN) module in the large language model since experts of FFN can store diverse task knowledge~\citep{geva2020transformer} from multi-task learning.
In the popular LLaMA-style FFN, the input $x$ is computed using a SLU gate:
\begin{equation}
    x=W_d(W_u x\cdot\mathrm{silu}(W_g x))
\end{equation}
where $W_g\in\mathbf{R}^{d'\times d},W_u\in\mathbf{R}^{d'\times d},W_d\in\mathbf{R}^{d\times d'}$ and $d'=8/3d$.
In the LoRA forward pass, each linear layer $W\in\{W_d,W_u,W_g\}$ is updated using the LoRA module described in Eq.~\ref{lora}.
For MING-MOE, we replace the original LoRA module with our Mixture-of-Low-rank Adaptation~(MoLoRA) module and only update it during training.
The MoLoRA FFN is shown in Figure~\ref{fig:molora_linear}.
Consider an MoLoRA module with $N$ experts and each expert is denoted as $\{E_i\}_{i=1}^N$, the forward pass of the linear layer is formulated as:
\begin{equation}
    h=Wx+\Delta W x=Wx + \sum_{i=1}^K G(x)_i E_{i}(x)
\end{equation}
where $G(\cdot)=\mathrm{softmax}(W_r x),W_r\in\mathbb{R}^{N\times d}$ is the router in MoLoRA and $K$ is the top-$K$ selection in the router which holds the top-$K$ affinity with respect to current input $x$.
This automatic router computes the weight of each expert output under a given number of $K$ activated experts to achieve both efficacy and efficiency during the inference phase.
Assuming each LoRA expert shares the same rank $r$ and $\alpha$ and we obtain the weight $w_i$ from the router $G(\cdot)$, the overall output of the MoLoRA linear layer can be formulated as:
\begin{equation}
    h=Wx+\frac{\alpha}{r}\sum_{i=1}^K w_i \cdot B_i A_i x
\end{equation}
where $W$ inherits from the pre-trained linear weight and is frozen during training.

\subsection{Medical Fine-tuning Corpus}
We primarily collected data from four segments, totaling 300k training samples, to fine-tune MING-MOE. For the part of the medical nlp task,We collect 70k training data provided by promptCBLUE~\footnote{\url{https://github.com/michael-wzhu/PromptCBLUE}}, which contain 16 types of medical nlp tasks, including named entity recognition, entity relationship extraction, clinical terminology noramlization, event extraction and so on. For the medical license exams, we collect 130k multiple choice question with rational in total. The training data is composed of the training set of the MedMCQA~\citep{pal2022medmcqa}, CMExam~\citep{liu2024benchmarking}, the English and the Chinese parts of the MMedBench~\citep{qiu2024towards}. For the part of the medical question answering and medical dialogue, we collect data from MING~\citep{MING}, HuatuoGPT~\citep{wang2023huatuo} and HuatuoGPT-II~\citep{chen2023huatuogpt}, and randomly select 100k sample from the collected data.

%% file: 4_exp.tex
\begin{table}[]
\renewcommand{\arraystretch}{1.2}
\centering
\resizebox{\textwidth}{!}{%
\begin{tabular}{ccccc}
\toprule
\textbf{Type} & \textbf{Task} & \textbf{Description}                                                   & \textbf{Size}                & \textbf{Metrics} \\
\midrule
              & CMeEE         & Chinese Medical Named Entity Recognition                               & 5000                         & Micro-F1         \\
              & CMeIE         & Chinese Medical Text Entity Relationship Extraction                    & {\color[HTML]{595959} 3,585} & Micro-F1         \\
              & CHIP-CDN      & Clinical Terminology Normalization                                     & 2,000                        & Micro-F1         \\
              & CHIP-CDEE     & Clinical Discovery Event Extraction                                    & 384                          & Micro-F1         \\
              & IMCS-V2-NER   & Intelligent Medical Conversation System Named Entity Recognition       & 833                          & Micro-F1         \\
 &
  CHIP-MDCFNPC &
  Medical Dialog Clinical Findings Positive and Negative Classification &
  1,000 &
  Micro-F1 \\
              & IMCS-V2-SR    & Intelligent Medical Conversation System Symptom Recognition            & 833                          & Micro-F1         \\
              & IMCS-V2-DAC   & Intelligent Medical Conversation System Dialogue Action Classification & 833                          & Macro-F1         \\
              & IMCS-V2-MRG   & Intelligent Medical Conversation System Medical Report Generation      & 833                          & RougeL           \\
              & CHIP-CTC      & Clinical Trial Criterion                                               & 7,682                        & Micro-F1         \\
              & CHIP-STS      & Semantic Textual Similarity                                            & 4,000                        & Micro-F1         \\
              & KUAKE-IR      & Information Retrieval                                                  & 1,000                        & Micro-F1         \\
              & KUAKE-QIC     & Query Intent Criterion                                                 & 1,955                        & Micro-F1         \\
              & KUAKE-QQR     & Query Query Relevance                                                  & 1,600                        & Micro-F1         \\
              & KUAKE-QTR     & Query Title Relevance                                                  & 2,913                        & Micro-F1         \\
\multirow{-16}{*}{\begin{tabular}[c]{@{}c@{}}Medical \\ NLP Tasks\end{tabular}} &
  MedDG &
  Medical Dialog Generation &
  2,747 &
  RougeL \\
\midrule
              & MedQA         & United States Medical License Exams (USMLE)                & 1,273         & Accuracy         \\
 &
  MedMCQA &
  High-Quality AIIMS \& NEET PG entrance exam MCQs &
  {\color[HTML]{595959} 4,183} &
  Accuracy \\
              & CMB           & Comprehensive Multi-level Assessment for Medical Knowledge & 11,200        & Accuracy         \\
              & CMExam        & Chinese National Medical Licensing Examination             & 6,811         & Accuracy         \\
              & MMLU$^\dagger$          & Massive Multitask Language Understanding                   & 1,561         & Accuracy         \\
              & CMMLU$^\dagger$        & Chinese Massive Multitask Language Understanding           & 1,354         & Accuracy         \\
              & CEval$^\dagger$         & A Multi-Level Multi-Discipline Chinese Evaluation          & 41            & Accuracy         \\
\multirow{-8}{*}{\begin{tabular}[c]{@{}c@{}}Medical \\ License Exams\end{tabular}} &
  CNPLE &
  2023 Chinese National Pharmacist Licensure Examination &
  960 &
  Accuracy \\
\bottomrule
\end{tabular}%
}
\caption{Statistics of the evaluated medical tasks. "$\dagger$" indicates that we only choose the questions related to the medical domain for the task.}
\label{tab:statistics}
\end{table}

\section{Experiments}
\subsection{Base Model and Configuration}
MING-MOE is a bilingual (Chinese and English) medical large language model built upon the foundation of the Qwen1.5-Chat~\footnote{\url{https://github.com/QwenLM/Qwen1.5}} model. Given the robust instruct-following capability of Qwen1.5-Chat, the fine-tuning process primarily focused on enhancing the model's ability to handle specific tasks in the medical domain. Meanwhile, a certain proportion of medical question-answering and interaction data was retained in the fine-tuning dataset to preserve the model's generative capabilities. Based on the size of the base model, we proposed four different sizes of MING-MOE, which include MING-MOE~(1.8B), MING-MOE~(4B), MING-MOE~(7B), and MING-MOE~(14B).

\subsection{Implementation Detail}
For all sizes of MING-MOE models, We set the batch size to 32 and train the models with 1 epoch. The learning rate is set to 2e-4, with the $warmup\_ratio=0.03$ and the cosine learning schedule. The maximum length of the training sample is configured to 3072. For the configuration of the PEFT, the LoRA rank $r$ and $\alpha$ are fixed at 16 and 32, respectively. The number of experts is set to 8 and 2 experts will be activated for each token during the training and inference. We only adopt MoLoRA for the multilayer perceptron~(MLP) layers in the feed-forward network~(FFN) blocks and adopt normal LoRA for the MLP layers in the attention blocks. All the experiments are conducted on 8$\times$A100 80G GPUs.

\subsection{Baseline Models}
We selected a diverse array of models to serve as baselines for comparing their performance on medical tasks. For the general open-source models, we choose Baichuan2-7b/13b-Chat~\citep{yang2023baichuan}, Qwen-7B/14B-Chat~\citep{bai2023qwen}, ChatGLM2/3-6b~\citep{zeng2022glm}, and Llama2-7B/13B-Chat~\citep{touvron2023llama}. For the medical open-source models, we choose DISC-MedLLM~\citep{bao2023disc}, HuatuoGPT~\citep{wang2023huatuo} and HuatuoGPT-II (7B/13B)~\citep{chen2023huatuogpt}. Additionaly, we also choose some closed-source model with strong performance, including ERNIE~\citep{sun2021ernie}, ChatGPT~\citep{elmohamed} and GPT4~\citep{DBLP:journals/corr/abs-2303-08774}.

\subsection{Medical Evaluation Benchmarks}
\paragraph{Medical NLP Tasks} The characteristic of Medical NLP Tasks lies in their unique input and output format requirements, primarily evaluating the models' ability to process medical texts and their alignment capabilities. Typically, only specialized models are capable of performing these tasks. CBLUE~\citep{zhang-etal-2022-cblue} is a Chinese multi-task medical dataset, encompassing 16 distinct medical tasks. Considering the highly varied input-output structures of these tasks, we adopted PromptCBLUE~\citep{zhu2023promptcblue}, which transforms CBLUE into a pure text format using specific prompts, ensuring compatibility with Large Language Models (LLMs). 

\paragraph{Medical Licensing Exams} Medical licensing exams are used to measure the medical knowledge and reasoning capabilities of large language models, serving as a common testing method. For this type of Benchmarks, we follow the experiments setting of~\citet{chen2023huatuogpt}. The examination benchmarks include: the US test set of  MedQA~\citep{jin2021disease}, the development set of MedMCQA~\citep{pal2022medmcqa}, and two comprehensive Chinese medical exam datasets, CMB~\citep{wang2023cmb} and CMExam~\citep{liu2024benchmarking}. We also collect the medical parts of the general benchmarks, which include: MMLU~\citep{hendrycks2020measuring}, C-Eval~\citep{huang2023ceval}, and CMMLU~\citep{li2023cmmlu}. We also test the performance of the models on the exam questions from the
2023 Chinese National Pharmacist Licensure Examination, which is collected by~\citet{chen2023huatuogpt}.



The statistics of the evaluated medical tasks are shown in Table~\ref{tab:statistics}.

\begin{table}[]
\renewcommand{\arraystretch}{1.2} 
\centering
\resizebox{\textwidth}{!}{%
\begin{tabular}{ccccccccc}
\toprule
\multirow{2}{*}{\textbf{Task}} &
  \multirow{2}{*}{\textbf{GPT4$^\dagger$}} &
  \multirow{2}{*}{\textbf{ChatGPT$^\dagger$}} &
  \multirow{2}{*}{\textbf{MOELoRA}} &
  \multirow{2}{*}{\textbf{HuatuoGPT-I}} &
  \multicolumn{4}{c}{\textbf{MING-MOE}} \\
             &       &                &                &       & \textbf{1.8B} & \textbf{4B} & \textbf{7B} & \textbf{13B} \\
\midrule
CMeEE        & 27.90 & -              & -              & -     & 61.67         & 67.29       & 69.64       & \textbf{71.37}        \\
CMeIE        & 8.30  & 30.58          & 51.93          & 18.26 & 43.56         & 47.50       & 56.75       & \textbf{57.13}        \\
CHIP-CDN     & 36.90 & 60.69          & 89.28 & 36.10 & 82.43         & 85.11       & 88.84       & \textbf{89.92}        \\
CHIP-CDEE    & 5.60  & 28.38          & 56.97          & 16.58 & 56.86         & 62.78       & 65.68       & \textbf{67.55}        \\
IMCS-V2-NER  & -     & -              & -              & -     & 83.75         & 86.87       & 86.95       & \textbf{89.67}        \\
CHIP-MDCFNPC & -     & 58.54          & 79.33          & 34.87 & 80.97         & 81.38       & 82.50       & \textbf{82.92}        \\
IMCS-V2-SR   & -     & -              & -              & -     & 66.67         & 70.03       & \textbf{70.35}       & 70.19        \\
IMCS-V2-DAC  & -     & -              & -              & -     & 5.57          & 6.06        & \textbf{6.17}        & 5.03         \\
IMCS-V2-MRG*  & 29.00 & 32.53          & 36.81          & 24.01 & 44.21         & 45.68       & 45.92       & \textbf{46.52}        \\
CHIP-CTC     & 55.00 & 52.53          & 86.91          & 19.09 & 79.80         & 83.78       & \textbf{88.17}       & 87.16        \\
CHIP-STS     & -     & -              & -              & -     & 77.33         & 78.64       & 79.98       & \textbf{82.49}        \\
KUAKE-IR     & -     & -              & -              & -     & 74.30         & 82.51       & 81.91       & \textbf{84.68}        \\
KUAKE-QIC    & -     & 48.51          & 86.75          & 14.54 & 76.11         & 84.87       & 86.62       & \textbf{87.00}        \\
KUAKE-QQR    & -     & -              & -              & -     & 70.56         & 73.99       & \textbf{77.78}       & 77.47        \\
KUAKE-QTR    & -     & -              & -              & -     & 56.06         & 62.08       & 64.20       & \textbf{67.29}        \\
MedDG*        & 10.90 & \textbf{13.61} & 10.89          & 13.08 & 11.18         & 11.37       & 12.28       & 12.95 \\
\midrule
Avg. & - & - & - & - & 60.69 & 64.37 & 66.48 & \textbf{67.46} \\
\bottomrule
\end{tabular}%
}
\caption{The results of medical large language models on each medical NLP task. Each task's score is calculated based on the metrics corresponding to that task in Table~\ref{tab:statistics}. The rows labeled with ``Avg.'' represent the average performance of each model on medical NLP tasks. Note that the models with "$\dagger$" indicate the results obtained from the OpenCompass~\citep{2023opencompass}. "MOELoRA" indicates the method proposed by~\citet{liu2023moelora}}
\label{tab:medical_nlp_tasks}
\end{table}

\subsection{Results on Medical NLP Tasks}
The results of the medical NLP tasks are shown in Table~\ref{tab:medical_nlp_tasks}. It can be observed that except for MedDG where ChatGPT achieved the highest score, MING-MOE has delivered the best results in all other tasks. Additionally, we found that MING-MOE (7B) has surpassed MING-MOE (14B) in some tasks and has overall achieved a comparable level of performance. This suggests that medical NLP tasks do not demand a high level of medical knowledge; rather, they require the model to handle various inputs and generate outputs in the appropriate formats.

\begin{table}[]
\renewcommand{\arraystretch}{1.2}
\centering
\resizebox{\textwidth}{!}{%
\begin{tabular}{ccccccccc}
\toprule
\textbf{Model} & \textbf{MedQA} & \textbf{MedMCQA} & \textbf{CMB} & \textbf{CMExam} & \textbf{MMLU} & \textbf{CMMLU} & \textbf{CEval} & \textbf{AVERAGE} \\
\midrule
DISC-MedLLM        & 28.67 & -     & 32.47 & 36.62 & -     & -     & -     & 32.59    \\
HuatuoGPT          & 25.77 & 31.2  & 28.81 & 31.07 & 34.91 & 33.23 & 36.53 & 31.65    \\
Llama2-7B-Chat     & 30.79 & 37.32 & 24.81 & 25.05 & 45.68 & 31.02 & 26.4  & 31.58    \\
Llama2-13B-Chat    & 36.68 & 39.66 & 26.51 & 26.6  & 50.67 & 32.64 & 31.47 & 34.89    \\
ChatGLM2-6B        & 24.98 & 33.33 & 42.41 & 43.55 & 40.94 & 43.87 & 47.20 & 39.47    \\
ChatGLM3-6B        & 28.75 & 35.91 & 39.81 & 43.21 & 47.21 & 46.97 & 48.80 & 41.52    \\
Baichuan2-7B-Chat  & 33.31 & 38.90 & 46.33 & 50.48 & 50.29 & 50.74 & 51.47 & 45.93    \\
Baichuan2-13B-Chat & 39.43 & 41.86 & 50.87 & 54.90 & 56.31 & 52.95 & 58.67 & 50.71    \\
Qwen-7B-Chat       & 33.46 & 41.36 & 49.39 & 53.33 & 53.88 & 54.65 & 52.80 & 48.41    \\
Qwen-14B-Chat      & 42.81 & 46.59 & 60.28 & 63.57 & 61.69 & 64.55 & 65.07 & 57.79    \\
ChatGPT(API)       & \textbf{52.24} & \textbf{53.60} & 43.26 & 46.51 & \textbf{69.96} & 50.37 & 48.80 & 52.11    \\
HuatuoGPT-II(7B)   & 41.13 & 41.87 & 60.39 & 65.81 & 51.44 & 59.08 & 62.40 & 54.59    \\
HuatuoGPT-II(13B)  & 45.68 & 47.41 & 63.34 & 68.98 & 54.00 & 61.45 & 64.00 & 57.84    \\
\midrule
MING-MOE (1.8B)    & 30.01 & 32.03 & 43.74 & 49.68 & 38.31 & 48.52 & 48.78 & 41.58    \\
MING-MOE (4B)      & 38.81 & 35.45 & 51.18 & 58.07 & 48.56 & 54.28 & 65.85 & 50.31    \\
MING-MOE (7B) & 47.92 & 39.25 & 62.14 & 69.18 & 57.21 & 64.99 & 58.54 & 57.03 \\
MING-MOE (14B)     & 51.85 & 41.21 & \textbf{70.96} & \textbf{76.19} & 62.08 & \textbf{71.79} & \textbf{68.29} & \textbf{63.20}  \\
\bottomrule
\end{tabular}%
}
\caption{The results on medical benchmark. Evaluation was done using validation data for MedQA, MedMCQA, and CMB. signifies extraction of only medical-related questions. ’-’ indicate that
the model cannot follow question and make a choice. Note that for the general benchmarks MMLU, CMMLU and CEval, we only choose the question related to the medical domain. The results of the baseline models are obtained from~\citet{chen2023huatuogpt}}
\label{tab:medical benchmark}
\end{table}

\begin{table}[]
\renewcommand{\arraystretch}{1.2}
\centering
\resizebox{\textwidth}{!}{%
\begin{tabular}{cccccccccccc}
\toprule
\multirow{2}{*}{\textbf{MODEL}} &
  \multicolumn{5}{c}{\textbf{Pharmacist Licensure Examination (Pharmacy)}} &
  \multicolumn{5}{c}{\textbf{Pharmacist Licensure Examination (TCM)}} &
  \multirow{2}{*}{\textbf{AVERAGE}} \\
 &
  \textbf{\begin{tabular}[c]{@{}c@{}}Optimal\\ Choice\end{tabular}} &
  \textbf{\begin{tabular}[c]{@{}c@{}}Matched\\ Selection\end{tabular}} &
  \textbf{\begin{tabular}[c]{@{}c@{}}Integrated\\ Analysis\end{tabular}} &
  \textbf{\begin{tabular}[c]{@{}c@{}}Multiple\\ Choice\end{tabular}} &
  \textbf{TotalScore} &
  \textbf{\begin{tabular}[c]{@{}c@{}}Optimal\\ Choice\end{tabular}} &
  \textbf{\begin{tabular}[c]{@{}c@{}}Matched\\ Selection\end{tabular}} &
  \textbf{\begin{tabular}[c]{@{}c@{}}Integrated\\ Analysis\end{tabular}} &
  \textbf{\begin{tabular}[c]{@{}c@{}}Multiple\\ Choice\end{tabular}} &
  \textbf{TotalScore} &
   \\
\midrule
DISC-MedLLM        & 22.20 & 26.80 & 23.30 & 0.00  & 22.60 & 24.40 & 32.30 & 15.00 & 0.00  & 24.90 & 23.80 \\
HuatuoGPT          & 25.60 & 25.50 & 23.30 & 2.60  & 23.40 & 24.10 & 26.80 & 31.60 & 7.50  & 24.90 & 24.20 \\
ChatGLM2-6B        & 37.00 & 36.80 & 25.00 & \textbf{31.70} & 35.30 & 33.10 & 37.30 & 35.00 & 37.30 & 33.70 & 34.50 \\
ChatGLM3-6B        & 39.50 & 39.10 & 10.50 & 0.20  & 34.60 & 31.80 & 38.20 & 25.00 & 20.00 & 32.90 & 33.80 \\
Qwen-7B-chat       & 43.80 & 46.80 & 33.30 & 18.40 & 41.90 & 40.00 & 43.20 & 33.30 & 17.50 & 38.80 & 40.40 \\
Qwen-14B-chat      & 56.20 & 58.60 & 41.70 & 21.10 & 52.70 & 51.30 & 51.00 & 27.50 & \textbf{41.70} & 47.90 & 50.30 \\
Biachuan2-7B-Chat  & 51.20 & 50.90 & 30.00 & 2.60  & 44.60 & 48.10 & 46.00 & 35.00 & 7.50  & 42.10 & 43.40 \\
Biachuan2-13B-Chat & 43.80 & 52.70 & 36.70 & 7.90  & 44.20 & 41.30 & 46.40 & 43.30 & 15.00 & 41.70 & 43.00 \\
ERNIE               & 45.00 & 60.90 & 36.70 & 23.70 & 49.60 & 53.80 & 59.10 & 38.30 & 20.00 & 51.50 & 50.60 \\
ChatGPT~(API)       & 45.60 & 44.10 & 36.70 & 13.20 & 41.20 & 34.40 & 32.30 & 30.00 & 15.00 & 31.20 & 36.20 \\
GPT-4~(API)         & 65.10 & 59.60 & \textbf{46.70} & 15.80 & 57.30 & 40.60 & 42.70 & 33.30 & 17.50 & 38.80 & 48.10 \\
HuatuoGPT-II (7B)  & 41.90 & 61.00 & 35.00 & 15.70 & 47.70 & 52.50 & 51.40 & 41.70 & 15.00 & 47.50 & 47.60 \\
HuatuoGPT-II (13B) & 47.50 & 64.10 & 45.00 & 23.70 & 52.90 & 48.80 & 61.80 & 45.00 & 17.50 & 51.60 & 52.30 \\
\midrule
MING-MOE (1.8B)    & 40.63 & 42.73 & 26.67 & 12.50 & 37.50 & 35.00 & 40.91 & 25.00 & 5.00  & 33.96 & 35.73 \\
MING-MOE (4B)      & 45.63 & 47.73 & 31.67 & 7.50 & 41.67 & 51.88 & 46.82 & 45.00 & 7.50 & 45.00 & 43.33 \\
MING-MOE (7B)      & 51.88 & 61.36 & 33.33 & 15.00 & 50.83 & 53.13 & 52.73 & 46.67 & 20.00 & 49.58 & 50.21 \\
MING-MOE (14B)     & \textbf{65.00} & \textbf{67.27} & 45.00 & 30.00 & \textbf{60.63} & \textbf{65.63} & \textbf{62.27} & \textbf{56.67} & 27.50 & \textbf{59.79} & \textbf{60.21} \\
\bottomrule
\end{tabular}%
}
\caption{Results of the 2023 Chinese National Pharmacist Licensure Examination. It consists of
two separate Examinations including Pharmacy track and Traditional Chinese Medicine (TCM)
Pharmacy track. The results of the baseline models are obtained from~\citet{chen2023huatuogpt}}
\label{tab:cnple}
\end{table}

\subsection{Results on Medical License Exams}
We tested a total of seven datasets to comprehensively evaluate the medical knowledge level of our medical model. As shown in Table~\ref{tab:medical benchmark}, MING-MOE has reached state-of-the-art levels in medical knowledge, particularly MING-MOE (14B), which even surpasses ChatGPT and the best-performing open-source medical model, HuatuoGPT-II. We observed that MING-MOE underperforms on three English datasets—MedQA, MedMCQA, and MMLU—which may be attributed to the base model Qwen's relatively weaker capabilities in English. Despite these challenges, MING-MOE still achieves impressive overall performance. Notably, MING-MOE (1.8B) is able to match the performance of ChatGLM3-6b, MING-MOE (4B) achieves similar results to Baichuan2-13b, and MING-MOE (7B) reaches performance levels comparable to HuatuoGPT-II (13B). This demonstrates that MING-MOE utilizes medical knowledge more efficiently than typical models, achieving higher levels of medical expertise with fewer parameters and effectively advancing the progress towards lighter large language models in the medical field.

Furthermore, we also tested the performance of our model on the 2023 Chinese National Medical Licensing Examinations. This dataset, due to its recency, is not susceptible to leakage through online data, allowing for a fairer assessment of the capabilities of medical large language models. As shown in Table~\ref{tab:cnple}, MING-MOE continued to achieve the best results. MING-MOE (7B) and MING-MOE (13B) even surpassed GPT-4. This sufficiently demonstrates MING-MOE's strong generalization capabilities in the medical field, making it one of the best models in terms of knowledge and reasoning abilities among all currently available open-source medical models.

\subsection{Case Study}
\begin{figure}[tbp]
    \centering
    \includegraphics[width=.95\linewidth]{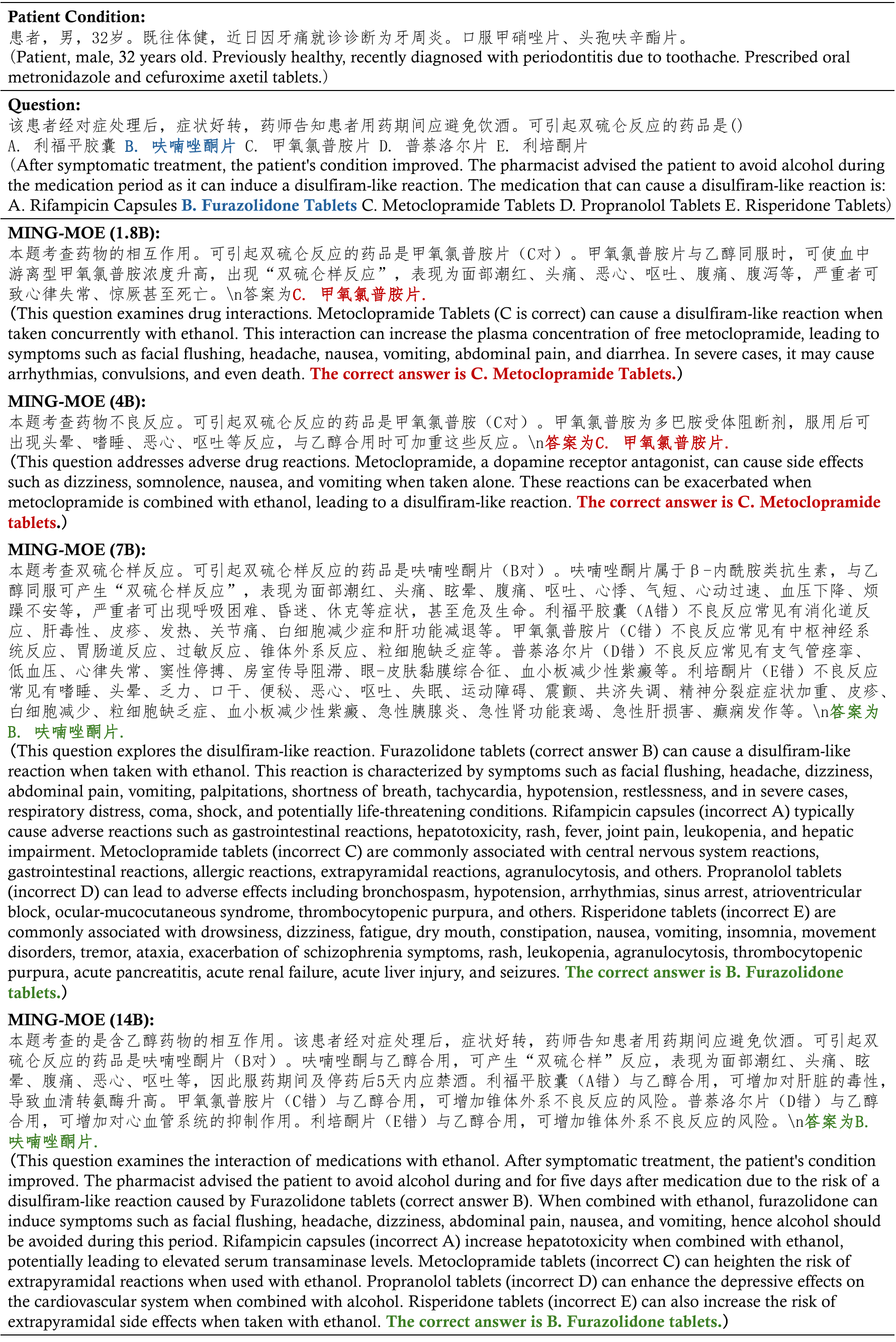}
    \caption{Case study on the sample from Chinese National Pharmacist Licensure Examination Pharmacy track. The font with blue color in the question indicates the correct options, the font with red color indicates the wrong answer of the models and the font with green color indicates the correct answer of the models.}
    \label{fig:case1}
\end{figure}

\begin{figure}[tbp]
    \centering
    \includegraphics[width=\linewidth]{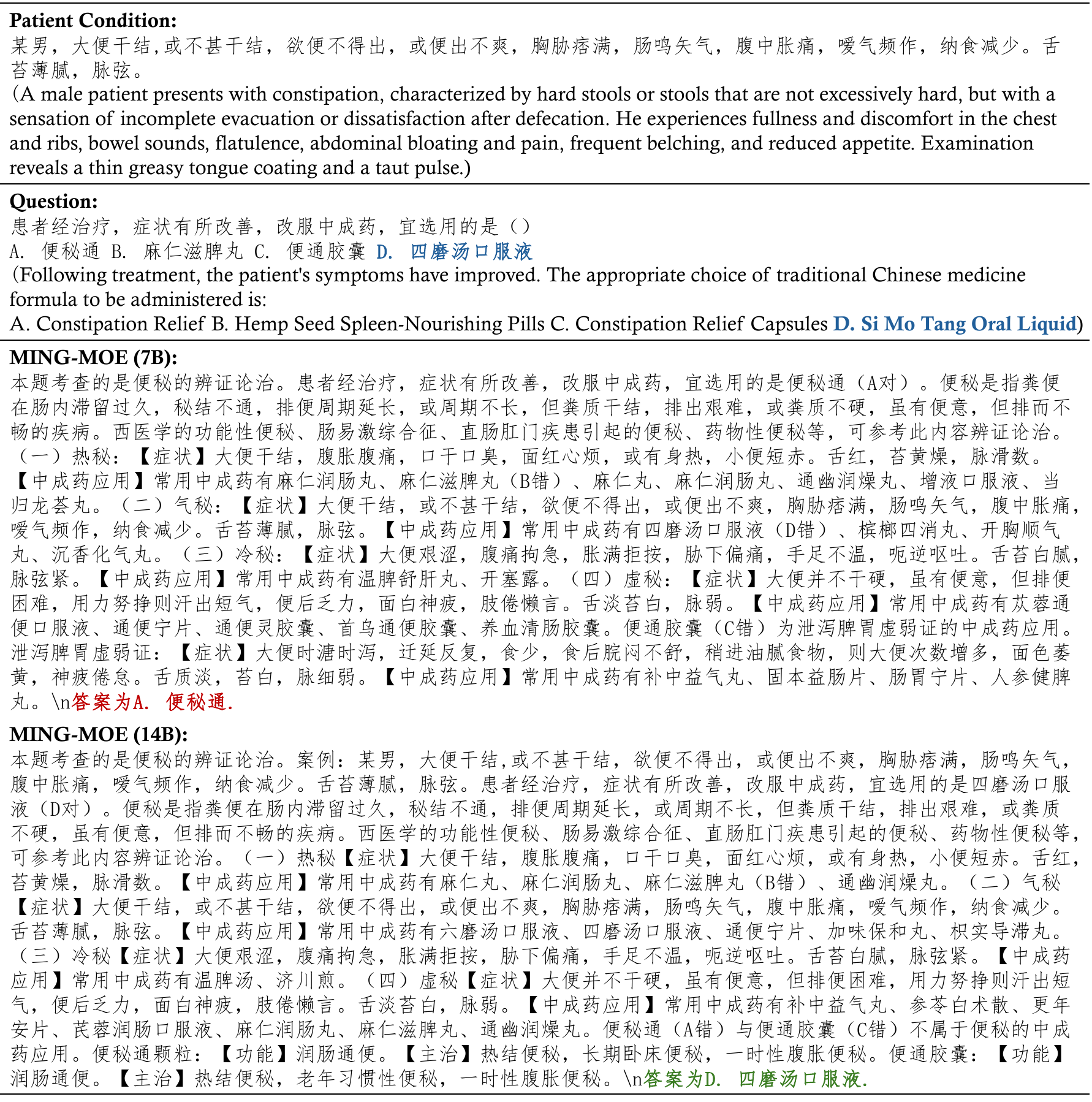}
    \caption{Case study on the sample from Chinese National Pharmacist Licensure Examination Traditional Chinese Medicine track. The font with blue color in the question indicates the correct options, the font with red color indicates the wrong answer of the models and the font with green color indicates the correct answer of the models.}
    \label{fig:case2}
\end{figure}

To gain an intuitive understanding of the performance of MING-MOE, we conducted case studies on different sizes of MING-MOE. Figures~\ref{fig:case1} and~\ref{fig:case2} illustrate examples randomly sampled from two tracks of the Chinese National Pharmacist Licensure Examination. It is observable that the model not only provides simple answers to examination questions but also offers relevant information and principles, enhancing the credibility and interpretability of the model outputs.

%% file: 5_discussion.tex

%% file: 6_conclusion.tex
\section{Conclusion}
In this paper, we introduce a novel MOE-based medical large language model, MING-MOE. By employing an Mixture of Low-Rank Adaptation~(MoLoRA) technique to enhance the multi-task learning of the base model, we found that MING-MOE achieves state-of-the-art performance across a variety of medical tasks. Additionally, compared to other models, MING-MOE can store more medical knowledge with fewer parameters and surpass existing medical models in various physician licensing examinations, even outperforming ChatGPT and GPT-4 in some respect.